\documentclass{article}

\usepackage{arxiv}

\usepackage[utf8]{inputenc} % allow utf-8 input
\usepackage[T1]{fontenc}    % use 8-bit T1 fonts
\usepackage{hyperref}       % hyperlinks
\usepackage{url}            % simple URL typesetting
\usepackage{booktabs}       % professional-quality tables
\usepackage{amsfonts}       % blackboard math symbols
\usepackage{nicefrac}       % compact symbols for 1/2, etc.
\usepackage{microtype}      % microtypography
\usepackage{lipsum}
\usepackage{multirow}
\usepackage{graphicx}
\usepackage{listings}
\usepackage{nicefrac}       % compact symbols for 1/2, etc.
\usepackage{microtype}      % microtypography
\usepackage{xcolor}
\usepackage{verbatim}
\usepackage{fontawesome}
\usepackage[utf8]{inputenc} % allow utf-8 input
\usepackage[T1]{fontenc}    % use 8-bit T1 fonts
\usepackage{hyperref}       % hyperlinks
\usepackage{booktabs}       % professional-quality tables
\usepackage{amsfonts}       % blackboard math symbols
\usepackage{nicefrac}       % compact symbols for 1/2, etc.
\usepackage{microtype}      % microtypography
\usepackage[english]{babel}
\usepackage{hyperref}
\hypersetup{
    colorlinks=true,
    linkcolor=blue,
    filecolor=magenta,      
    urlcolor=cyan,
}
\definecolor{codegreen}{rgb}{0,0.6,0}
\definecolor{codegray}{rgb}{0.5,0.5,0.5}
\definecolor{codepurple}{rgb}{0.58,0,0.82}
\definecolor{backcolour}{rgb}{0.95,0.95,0.92}
 
\lstdefinestyle{mystyle}{
    backgroundcolor=\color{backcolour},   
    commentstyle=\color{codegreen},
    keywordstyle=\color{magenta},
    numberstyle=\tiny\color{codegray},
    stringstyle=\color{codepurple},
    basicstyle=\ttfamily\footnotesize,
    breakatwhitespace=false,         
    breaklines=true,                 
    captionpos=b,                    
    keepspaces=true,                 
    numbers=left,                    
    numbersep=5pt,                  
    showspaces=false,                
    showstringspaces=false,
    showtabs=false,                  
    tabsize=2
}
\title{A framework for democratizing AI}

\author{
  Shakeel Ahmed, Ravi S. Mula, Soma S. Dhavala\thanks{corresponding author}\\
  \texttt{\{shakkeel, ravi, soma\}@mlsquare.org}
}

\begin{document}
\maketitle

\begin{abstract}
Machine Learning and Artificial Intelligence are considered an integral part of the Fourth Industrial Revolution. Their impact, and far-reaching consequences, while acknowledged, are yet to be comprehended. These technologies are very specialized, and few organizations and select highly trained professionals have the wherewithal, in terms of money, manpower, and might, to chart the future. However, concentration of power can lead to marginalization, causing severe inequalities. Regulatory agencies and governments across the globe are creating national policies, and laws around these technologies to protect the rights of the digital citizens, as well as to empower them. Even private, not-for-profit organizations are also contributing to democratizing the technologies by making them \emph{accessible} and \emph{affordable}. However, accessibility and affordability are all but a few of the facets of democratizing the field. Others include, but not limited to, \emph{portability}, \emph{explainability}, \emph{credibility}, \emph{fairness}, among others. As one can imagine, democratizing AI is a multi-faceted problem, and it requires advancements in science, technology and policy. At \texttt{mlsquare}, we are developing scientific tools in this space. Specifically, we introduce an opinionated, extensible, \texttt{Python} framework that provides a single point of interface to a variety of solutions in each of the categories mentioned above. We present the design details, APIs of the framework, reference implementations, road map for development, and guidelines for contributions.
\end{abstract}

\section{Introduction}
In the 21st century, we take it for granted how to harness the potential of electricity. It may seem incomprehensible that specialist engineers had to be hired to operate electric generators \emph{in situ} at homes just to run light bulbs in the early days. It required a coming together of great minds, ripe business opportunities, technological breakthroughs, and abundance of patience, to turn electricity into what we now consider a mere utility. While Machine Learning(ML) and Artificial Intelligence(AI) have been around for a long time, with roots in various disciplines such as a Statistics, Computer Science, Neuroscience, Physics, Philosophy, among others, the ecosystem seems to be ripe for deriving utilitarian value out of them. We are already witnessing their proven utility in Information Retrieval, and e-commerce, to name a few. But it is just the tip of the ice-berg. They are a part of a much larger, sweeping 4th Industrial Revolution\cite{Schwab}, whose ramifications are too significant to ignore, and too complex to fathom. As it stands today,
\renewcommand{\thefootnote}{\fnsymbol{footnote}}
AI\footnotemark[2] is a specialized field, and a select few have the expertise required to shape them for the benefit of the society at large. A plausible solution is to democratize the very creation, distribution, and consumption of the technology. A concerted cooperation between public and private entities in creating both scientific and technological tools is necessary. Governments are reacting by creating national policies and task forces \cite{niti}, and new laws are being formed to protect the rights of data citizens\cite{gdpr,futureai}. Many institutions, both academic and private\cite{mit,OpenAI,AllenAI, google,h2o} are addressing this problem by heavily investing in advancing the science and technology. Below, we take a cursory look at some desirable attributes of AI that make it responsible, and responsive. 

\begin{itemize}
  \item\emph{Portable: Separate the creation of AI from its consumption}\\ 
  \href{https://en.wikipedia.org/wiki/Java_(programming_language)}{\texttt{JAVA}} transformed the enterprise software by making it portable across platforms -- code once written, can be run everywhere.  Likewise, intelligence (encoded in the form of models) once created, shall run anywhere. Some existing approaches are via porting models in \texttt{PMML}\cite{PMML} format, or compile them to \texttt{ONNX}\cite{ONNX} format, implemented in \texttt{winML}\cite{WinML}. Another approach is to embrace \emph{write-once, scale-anyhow} paradigms like \texttt{Spark}\cite{spark} and \texttt{Julia}\cite{bezanson2017julia}. Our approach is discussed in  Section \ref{portability section}.
  
  \item\emph{Explainable\footnotemark[2]: Provide plausible explanations, along with predictions} \footnotetext[2]{AI \& ML, Explainability \& Interpretability, Models \& Algorithms are used interchangeably for the sake of simplicity}
 When decisions are made by AI agents in high stakes environments, such as those encountered in FinTech, EdTech, and Healthcare, it is imperative that a human being would be able to interpret those decisions in the context of the situation, and explain how the agent has come to that decision\cite{somadot,Rudin}. Tools such as \texttt{LIME}\cite{LIME}, \texttt{CORELS}\cite{CORELS}, \texttt{DoWhy}\cite{dowhy}, \texttt{shapely}\cite{shapely} are some practical approaches to providing explanations. For an overview see here\cite{molnar}. Our approach is outlined in  Section \ref{explainability section}.

  \item\emph{Credible: Provide coverage intervals, and other uncertainty quantification metrics} \\
  Statistical Software such as \href{https://SAS.com}{\texttt{SAS}\textsuperscript{\textregistered}},\href{https://www.stata.com/}{\texttt{STATA}\textsuperscript{\textregistered}}, and many libraries in \texttt{R}\cite{R}, provide uncertainty estimates, along with predictions, in terms of standard errors and p-values. Bayesian counterparts produce credible intervals. Statistics community lays enormous importance on producing such inference summaries. Unfortunately, ML tools rarely provide such reports, as they are primarily concerned with prediction tasks alone. This problem is exasperated in the Deep Learning space, where they reportedly can make confident mistakes\cite{nguyen2015deep}. How can we make AI credible by not only making predictions, but also by reporting coverage intervals and other uncertainty estimates around the predictions? For a curated list of papers, and tools, see \cite{uq-deep}.
  
  \item \emph{Fair: Make AI-bias free and equitable}\\ 
  It is said, what we eat is what we are and what we think is what we become. AI is no different. What is predicted is based on what data the AI systems are fed. As a result, AI systems can make biased, unethical decisions, putting some sections of the community at risk. How can AI be made fair and equitable? Recent, on going work by IBM, provides some alternatives to detecting bias, and acting on it \cite{aif360-oct-2018}.

  \item \emph{Decentralized and Distributable - Let models go where data is, instead of the other way round}\\
  All internet-first commercial players rely on user generated data to provide hyper personalized products. Although such products provide tangible benefits to the end users, they come at the cost of risking private data and security threats. A decentralized system wherein the data remains within the users' premises can potentially avoid such privacy issues. This pivoting has lead to the popularization of techniques such as Secure Multiparty Computation(SMPC) and Federated Learning. SMPC restricts exposing model weights, while Federated Learning takes care of securely distributing the training process to multiple worker nodes. Frameworks like PySyft\cite{Pysyft} makes SMPC and Federated Learning intuitive and accessible to machine learning developers. Decentralizing the distribution channel of data, and intelligence, can mitigate some risks of centralized control.
  
  \item \emph{Declarative: Say what the model needs and what it should do. Do not worry about how it should go about doing it}\\ 
  An important aspect of ML model development is getting the right data, and model management. Many practitioners estimate that more than 80\% of Developer time goes in the ETL (Extract, Transform, Load) jobs. However, Developers should only be concerned with \emph{what} they need and not \emph{how} they should get the data. This problem is largely addressed by the databases world with standards like \texttt{SQ}. Many databases, including NoSQL databases, support \texttt{SQL} for querying and retrieving the data. Project \texttt{Morpheous}\cite{cypher} certainly looks very interesting as it makes graph and tabular data interoperable seamlessly. It is a practical alternative to associative arrays that attempt to unify RDBS, Graphs, and NoSQL databases\cite{d4d}. That means that Data Scientist can leverage SQL/OpenCypher for asking what they need and offload the “getting” part to the database’s compute engines. While such open standards are not available for model specification, the workhorse \texttt{'lm'} package in \texttt{R} allows models to be specified in the formula syntax. And \texttt{SAS}'s language certainly looks like the SQL analogue for model specification.  Can we make SQL/OpenCypher the \emph{de facto} declarative language for ETL and  develop a standard to declaratively specify the entire pipeline, including model specification and data flow specification?  For an outline of the proposal, see here\cite{dagger}.
  
  \item \emph{Reproducible: Reproduce any result, on-demand}\\
  Model building and doing Data Science is experimental in nature. Unless carefully managed, reproducing the results is next to impossible, particularly when working in distributed environments. At the very least, all results shall be reproducible. It is possibly by ensuring that 1) Data 2) Models and 3) Run Time Environments are versioned. An open source ML-As-A-Service platform, \texttt{daggit}\cite{daggit}, promises to handle this responsibility by using \texttt{DVC}\cite{dvc}, \texttt{git}\cite{git} and \texttt{docker}\cite{docker}, for versioning Data, Models, and Run Times, respectively. 
  
\end{itemize}

The above list is neither comprehensive nor exhaustive. There are many other challenges to over come such as making AI systems \emph{scalable}, \emph{auditable}, \emph{actionable}, \emph{governable}, \emph{discoverable} among others. But, as it appears, solutions are fragmented, and a holistic viewpoint is missing. At \texttt{mlsquare}, we are developing a single point of interface to bring several solutions together. In the next Sections, we focus on the specific solutions we are building.

\section{Democratizing AI at \texttt{mlsquare}}

We provide an extensible \texttt{Python} framework that incorporates the above mentioned tools, where possible, and innovate where necessary. The following are the design goals:
\begin{itemize}
    \item \emph{Bring Your Own Spec First}. Use existing, well-documented APIs, such as those in \texttt{sklearn} for ML. Only provide implementations or extend the functionality where lacking.
    \item \emph{Bring Your Own Experience First}: Minimal changes shall be introduced in the development workflow. For example, with just one additional line of code to standard model fitting in \texttt{sklearn}, the resulting model can run on a GPU.
    \item \emph{Consistent}: All the quality attributes of AI described earlier shall become first class methods of a model object with a consistent interface. It should not be necessary to stitch different, possibly incompatible, functionalities. 
    \item \emph{Compositional}: Deep Learning, when looked from a technology perspective, is a lego block computational framework for composing models. They can support inputs and outputs of varying shape, size, and nature. It allows many models to be expressed compositionally, without having to implement every piece of it. The lego block architecture gives a lot of expressive power to the model designer. With backing from all hardware vendors, and framework developers, hardware acceleration is an added benefit.
    \item \emph{Modular}: Thanks to inherent modularity of Deep Learning technology, exploit the inherent object-orientedness of many algorithms. For example, when developing a Decision Tree equivalent in Deep Neural Network (DNN)\cite{DNDT}, write the Decision Tree as a DNN layer. Then, one could immediately use it either in a classification or regression task. Even one can develop a Kernel Decision Tree. This modularization amplifies the expressiveness.
    \item \emph{Extensible}: All the provided implementations shall be extensible or default behaviours can be overwritten if one chooses to.    
\end{itemize}
In the following subsections, we describe our approach to providing \emph{portability} and \emph{explainability}.
\subsection{Portability}\label{portability section}
As ML is practiced today, Data Scientist or an ML Scientist would write the models in a particular language (eg \texttt{Python}), and a Production Engineer would rewrite them in a different language or reimplement them in a different tech stack -- scalability being the primary concern. This is called as the two language problem. One way to deal with the two language problem is to unify the tech stack. Languages like \texttt{Julia} claim that same piece of code can scale from a laptop to a cluster. Frameworks like \texttt{Apache Spark} also support multiple languages, and scale from a single CPU system to a cluster. But a developer is tied to a single ecosystem. Another way to achieve portability is by means by having an intermediate representation of the models such as \texttt{PMML} and \texttt{ONNX}. Many Deep Learning frameworks such as a \texttt{TensorFlow}\cite{TensorFlow},\texttt{PyTorch}\cite{PyTorch} and \texttt{MXNet}\cite{MXNet} support \texttt{ONNX}. Of interest is \texttt{WinML} which allows saving \texttt{sklearn}, \texttt{xgboost}\cite{XgBoost} models in \texttt{ONNX}. It does so by reimplementing the models in a target language like \texttt{TensforFlow}. While such exact reimplementations might provide high fidelity portings, extending and supporting the models can be tedious and time consuming, if not impossible. Rather than providing such one-to-one operator level mappings of models, we suggest a more generic semantic mapping of models as an efficient alternative. We utilize three different approaches that enable this transpilation process, each with its own advantages and disadvantages. We focus on supervised tasks.

\begin{enumerate}

\item Exact Semantic Map\\
When an exact equivalence between a user supplied model(referred to as primal model), and it's neural network counterpart(referred to as proxy model) exists, we can initialize the weights of the neural network with the parameters of the fitted model, and just save the model. This can be achieved for many classical models that fall under the Generalized Linear Models umbrella. An advantage of this approach is that the proxy model is faithful to the primal model.

\item Approximate Semantic Map\\
In this approach, the proxy model is trained on same data as the primal model, but its target labels are the predictions from the primal model. Further, the proxy model's architecture is chosen so as to closely fulfill proxy model's intent. Together, they ensure a better semantic approximation. In someways, the capacity of the proxy model is constrained by the primal model's capacity.  In \textit{Section} \emph{cite results section}, we compare the the performance of such transpiled models. On the theoretical side, a Probably Approximately Faithful (PAF) framework has to be developed.

\item Universal Map\\
In the third approach, both the intent and implementation can be delegated to the proxy model. In someways, neural networks are used as generic black box functional approximators. This may be useful when the primal models can not be fit on large data sets, or there is no theoretical backing for a semantic map. Say, a user is interested in training a \texttt{sklearn Logistic Regression} model on a 1TB dataset. This would be extremely hard with existing \texttt{sklearn} implementations. On the other hand, a proxy model can be easily scaled with the underneath hardware. But the downside is that, fidelity between primal and proxy models may not be preserved. 
\end{enumerate}

\subsection{Explainability} \label{explainability section}
Explainability of black-box machine learning models is extremely critical while developing machine learning applications. Typically, explainability is focused on explaining what a model did or how a model was making predictions. While they are certainly useful for a Data Scientist, they are not really the explanations that an end-user needs. Instead, the explanations shall be about the predictions, and what an end user can do about them. We approach \emph{explainability} from this perspective. In particular, we define, at an abstract level, explanations as predicates in a First Order Logic system, represented in Conjunctive Normal Form (CNF). A proxy model shall produce such predicates as its output. In that light, we see producing explanations as a synthetic language generation problem.  

The \texttt{explain} method provides explanations of a model's output. \texttt{explain} is model agnostic and provides interpretations as a decision tree path. We use recurrent neural networks to train on the outputs of a localized decision tree for the given training dataset. When fed with a new unseen data point, this RNN would output the corresponding decision tree path traversed. This path provides us with the decision taken by the model at each feature and acts as an interpretation for the same. \texttt{explain} is an ongoing research and is only available as a prototype.

\section{Specifics of the Framework}

\begin{figure}[p]
\centering
\includegraphics[scale=0.6]{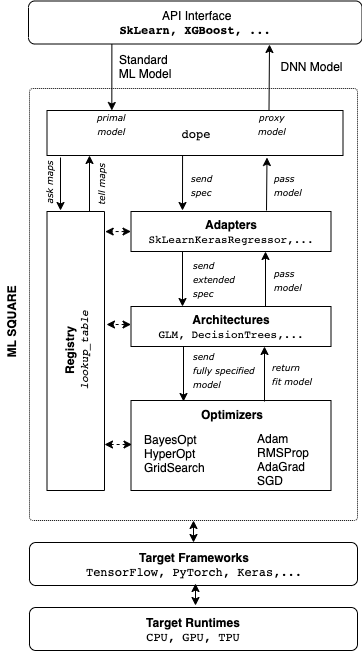}
\caption{Components in the Framework}
\label{components figure}
\end{figure}

The \texttt{mlsquare} framework utilizes multiple extensible components as shown in Fig.\ref{components figure}. 
Portability in the framework is achieved with a single line of additional code. The user is expected to pass their standard machine learning model(primal model) to the \texttt{dope} function. \texttt{dope} then handles the responsibility of detecting and triggering the corresponding neural network architecture with the help of \texttt{adapters} and \texttt{optimizers}. Internally, a neural network architecture search (NAS) occurs and  \texttt{dope} returns the optimal neural network model. The \texttt{Python} object returned is wrapped in such a way that it's interface remains identical to that of primal model sent to \texttt{dope}. Below, we outline the components that act as the foundation for \texttt{dope}'ing.

\subsection{Dope}
\begin{figure}[p]
 \centering
\includegraphics[scale=0.6]{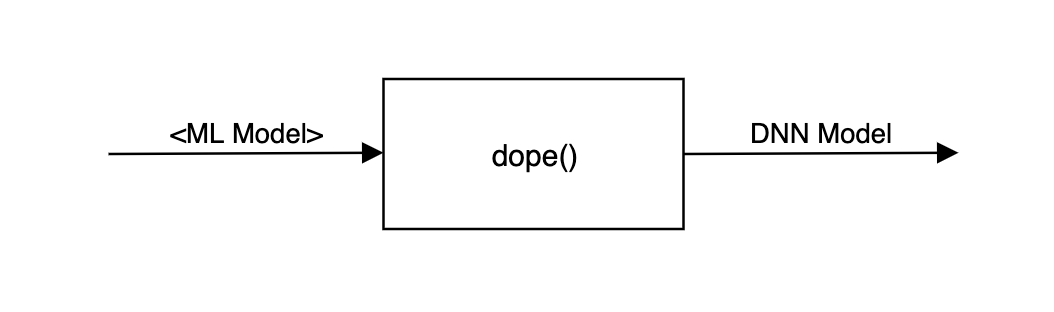}
\caption{Converting a Standard ML Model using \texttt{dope}}
\label{dope figure}
\end{figure}

\texttt{dope} accepts the primal model as an argument(see Fig. \ref{dope figure}). Additional model configuration can be passed as keyword arguments. This function handles the responsibility of mapping the primal model to its neural network equivalent. Once the right architectural parameters are obtained, \texttt{dope} delegates the responsibility of wrapping the neural network model with the characteristics of the primal model to \texttt{adapters}. The adapter returns a \texttt{Python} Object that behaves like the primal model but under the hood utilizes neural networks. As shown in Fig. \ref{components figure}, \texttt{dope} is assisted by two other components - \texttt{registry} and \texttt{adapters}.

\subsection{Adapter}

\texttt{Adapters} act as a connector between the primal model and the proxy model (i.e. the neural network equivalent). We currently support \texttt{Keras}\cite{Keras} as the backend for our proxy models and \texttt{sklearn} for primal models. \texttt{Adapters} provide methods(ex - \texttt{fit}, \texttt{score}, \texttt{predict} etc.) that extend on the proxy model’s APIs.  When the \texttt{fit} method is invoked, it triggers the \texttt{Optimizer}. The \texttt{Optimizer} receives model configurations and returns a trained model. \texttt{Adapter} then returns the trained model to \texttt{dope}. The \texttt{save} method exports the trained model as a \texttt{Keras} and \texttt{ONNX} models.
The \texttt{Adapter} module can hold multiple \texttt{Adapter} classes for each type of "primal-proxy" mappings. For example, we currently have two classes in \texttt{Adapters} - \texttt{SklearnKerasClassifier} and \texttt{SklearnKerasRegressor}. A \texttt{sklearn} classifier model, that needs to be mapped to a \texttt{Keras} model, can utilize the \texttt{SklearnKerasClassifier} adapter. New adapters can be written for interacting with other scientific libraries.

\subsection{Optimizer}
The neural network architecture of the proxy model can depend on the dataset it is being trained. Hence, we pass the proxy model through a neural architecture search (NAS) process. We use \texttt{Tune}\cite{Tune} based on \texttt{Ray}\cite{Ray} for our models’ NAS process. The optimizer component searches for best model and returns it to the adapter. Configuration of \texttt{tune} can be edited or reconfigured by passing the parameters via the \texttt{fit} method of the proxy model.

\subsection{Registry}

The \texttt{registry} object, like the name suggests, maintains a registry of models supported by ML Square. \texttt{registry} is a \texttt{python} dictionary object which is initialized while importing the \texttt{mlsquare} library. It expects a tuple of length two as key - name of the algorithm and the package name. \texttt{registry} then returns the corresponding proxy model and the adapter. \texttt{registry} can be used to register a model with the help of a decorator -- just add \texttt{@registry.register} decorator in the models' class definition. The model should contain adapter, model name and primal models' module name. Then, upon calling \texttt{dope}, a corresponding NAS can be triggered.

\subsection{Architectures}
\begin{figure}[p]
\centering
\includegraphics[scale=0.3]{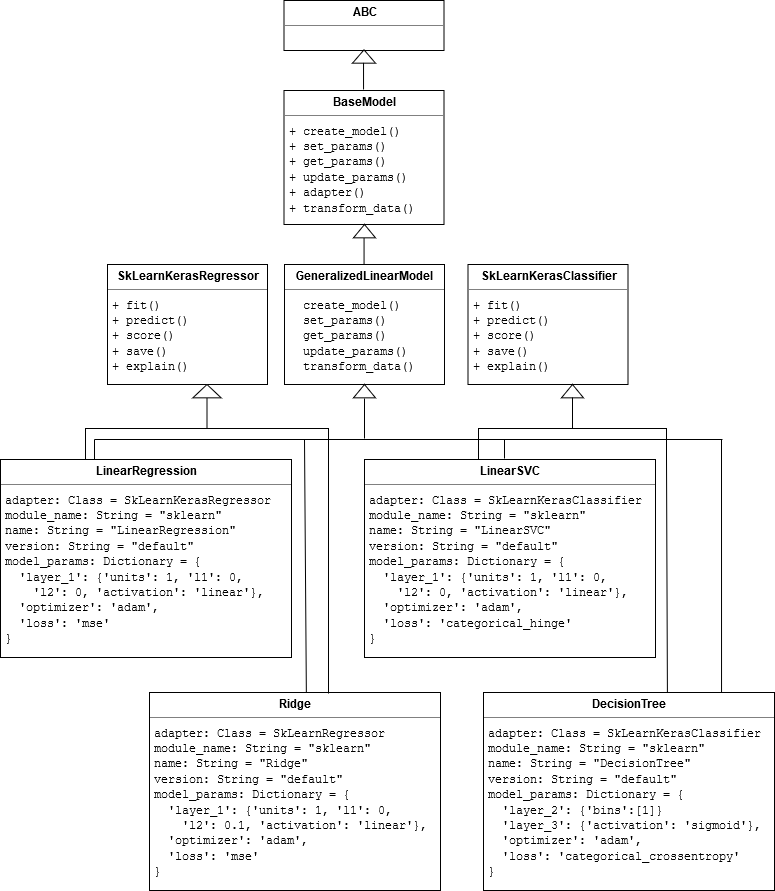}
\caption{UML Representation of \texttt{architecture}}
\label{ER diagram}
\end{figure}

The \texttt{architectures} module in \texttt{mlsquare} library provides the semantic mappings of the primal models. It is a \texttt{Python} \texttt{class} with various attributes to define the mappings. These attributes include the corresponding \texttt{adapter} of the model, module name of the \texttt{primal model}, name of the algorithm and model parameters required to initialize the neural network. To keep track of multiple versions of a given \texttt{proxy model}, an additional attribute called \texttt{version} is supported. There are three levels of class declarations involved in creating a model mapping. \texttt{BaseClass} is the topmost abstract base class (\texttt{ABC}). This class ensures that the basic methods expected for the functioning of a \texttt{proxy model} are not left undeclared. The next level is a generic parent class from which multiple algorithms can inherit common attributes. For example, the \texttt{GenralizedLinearModel} is used as a parent class for \texttt{LinearRegression}, \texttt{LogisticRegression}, \texttt{RidgeRegression} etc. The final level is declaring the \texttt{proxy model} itself as a class. This level includes declaring the key attributes required to create the model. \\
\newline
Taking a closer look at Fig. \ref{ER diagram} reveals how the modular aspects of neural networks are being utilized in \texttt{mlsquare} to accomplish semantic mappings. Tweaking the activation and loss function set in the model parameters of the \texttt{proxy model} that inherits \texttt{GeneralizedLinearModel} can provide a range of different models. It provides a lot of flexibility for the developers in deciding how a model mapping should be declared.

\section{Workflow}

\begin{figure}[hp]
 \centering
\includegraphics[scale=0.3]{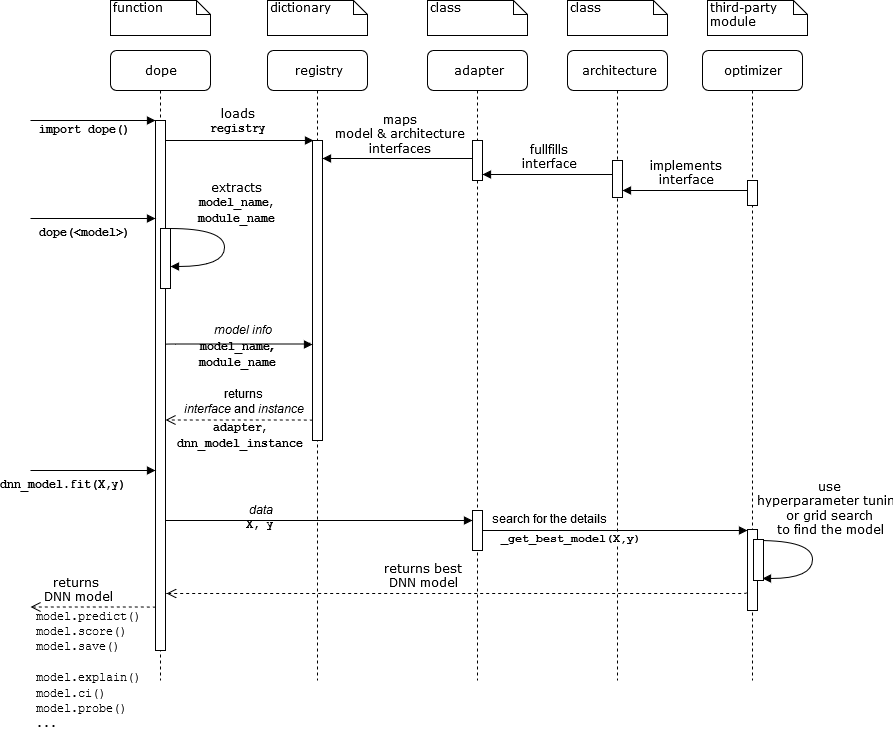}
\caption{Workflow representation}
\label{Workflow diagram}
\end{figure}

Figure \ref{Workflow diagram} shows a typical user workflow while utilizing \texttt{mlsquare}'s \texttt{dope} functionality. Such a workflow involves three touchpoints with \texttt{dope} while transpiling the model. 

\begin{enumerate}
    \item import the \texttt{dope} function
    \item pass the primal model to the \texttt{dope} function
    \item use the transpiled model to train, score, predict or save it for future use
\end{enumerate}

The following subsections provide details on the different methods available on the transpiled proxy model.
\subsection{Developer Experience}
The code snippet below demonstrates a typical workflow of training and scoring a \texttt{Sklearn} Linear Regression model.
\begin{lstlisting}[language=Python, caption=Usage of dope]
from mlsquare import dope
from sklearn.linear_model import LinearRegression
from sklearn.preprocessing import StandardScaler
from sklearn.model_selection import train_test_split
import pandas as pd
from sklearn.datasets import load_diabetes

model = LinearRegression()
diabetes = load_diabetes()

X = diabetes.data
sc = StandardScaler()
X = sc.fit_transform(X)
Y = diabetes.target

x_train, x_test, y_train, y_test = train_test_split(X, Y, test_size=0.60, random_state=0)

m = dope(model)

m.fit(x_train, y_train)
m.score(x_test, y_test)
\end{lstlisting}

 As mentioned earlier, adding one additional line of code(line no. 18) converts the \texttt{sklearn} model into a neural network. The object returned by \texttt{dope()} can then be used like any other \texttt{sklearn} model. The API interface remains consistent, with the only difference being that the training and scoring are performed on a neural network. The capabilities of this extended model are explained in detail in the following sections.

\subsubsection{\texttt{fit()} method}
Below shown is a code snippet for calling the \texttt{fit()} method.
\begin{lstlisting}[language=Python, caption=Customizing Neural Architecture Search]
# Train with default mappings
m.fit(x_train, y_train)

# Train with additional parameters
# tune by ray package is the NAS backend used by mlsquare
from ray import tune
# Define the hyperparameters you wish to customize
params = {'optimizer':{'grid_search':['adam', 'nadam']}}
m.fit(X_train, y_train, params=params)
\end{lstlisting}
When the \texttt{fit} method is called a neural architecture search is performed. If no additional parameters are given (line no.2), the default mappings are used to create and train the neural network. \texttt{dope} provides the option to customize the hyperparameters used for neural architecture search (line no. 8 and 9). In the above example, we customize the "optimizer" of the neural network. This, in turn, does a grid search on \texttt{adam} and \texttt{nadam} optimizers and returns the best performing model.

\subsubsection{\texttt{score()} and \texttt{predict()} methods}
The \texttt{dope}'d model contains both the \texttt{score} and \texttt{predict} methods like a \texttt{sklearn} model.
\begin{lstlisting}[language=Python, caption=Score and Predict methods]
# Scoring your models
m.score(x_test, y_test)
# Sample output
# 90/90 [==============================] - 0s 244us/step
# [0.5179819915029737, 0.911111111442248]

# Predict outputs
y_pred = m.predict(x_test)
\end{lstlisting}
 In the above example, the score method returns a list with two values - loss and accuracy. The predict method returns an array of predicted values.

\subsubsection{Saving dope'd model}
The save method available on the \texttt{dope}'d model allows exporting the trained model in two formats - \texttt{ONNX} and \texttt{HDF5}. The method expects a file name while saving the model.
\begin{lstlisting}[language=Python, caption=Saving a model]
# Save method expects file name as an argument
m.save(filename="my_model")
\end{lstlisting}
The ONNX file can then be loaded, converted or used directly in different runtimes. We could score a \texttt{sklearn} model in chrome browser. 

\section{Results}
\label{results section}
The framework currently contains a collection of seven \texttt{sklearn} models mapped to their neural network equivalents. Table \ref{results}  shows the performance comparison of each of these models. The comparisons are done by training and scoring each algorithm on a relevant dataset. The training and scoring are done on the standard \texttt{sklearn} model and its transpiled equivalent \texttt{dope}'d model.
\begin{table}[h!]
\centering
\begin{tabular}{ |p{3cm}||p{3cm}|p{3cm}|p{3cm}|p{3cm}|  }
 \hline
 \multicolumn{4}{|c|}{Performance comparison} \\
 \hline
 Algorithm& Dataset Name& Dope (acc/$r^2$)& sklearn (acc/$r^2$)\\
 \hline
 Logistic Regression   & Iris    &0.84&   0.82\\
 Linear Regression   & UCI Diabetes    &0.42&   0.40\\
 Ridge Regression   & UCI Diabetes    &0.44&   0.42\\
 Lasso Regression   & UCI Diabetes    &0.43&   0.43\\
 ElasticNet Regression   & UCI Diabetes    &0.44&   0.44\\
 Linear SVC   & Iris    &0.84&   0.91\\
 %SVC   & Iris    &0.46&   0.93\\
 Decision Tree Classifier   & Iris    &0.96&   0.93\\
 \hline
\end{tabular}
\caption{Comparison of \texttt{dope}'d models vs Standard \texttt{sklearn} models}
\label{results}
\end{table}

The datasets chosen were UCI Iris for regression algorithms and UCI Diabetes datasets\cite{UCI} for classification algorithms. The performance metric used was accuracy and $r^2$ score for classification and regression respectively. Table \ref{results} results clearly shows that the performance of models transpiled by \texttt{dope} are on par with their \texttt{sklearn} counterparts' performance.

A more exhaustive collection of trial runs and their corresponding results can be found in this \href{https://docs.google.com/spreadsheets/d/1E5jcq2w42gN8bMIaeaRJpAdhgSVN-2XDJ_YTHe4qfwY/edit?usp=sharing}{link}. This spreadsheet contains performance and configuration details of each new algorithm added to \texttt{mlsquare}. Once a new algorithm is added, a \texttt{Python} script trains and scores the algorithm with relevant datasets. By making it openly available to view the results, it helps us maintain accountability and quality of new algorithms added.

\section{Roadmap} \label{Roadmap section}
Table \ref{progress} depicts the available features[\checkmark], ongoing work[\small{\faFlask}] and future research[\small{\faSpinner}] on the \texttt{mlsquare} framework. One of the goals of building this framework is to propose and demonstrate the feasibility of a machine learning framework that can ease the process of democratizing AI, utilizing existing tools with very minimal changes. The \texttt{mlsquare} framework currently supports the portability feature(\texttt{.save()} method). It can convert above mentioned \texttt{sklearn} methods to a neural network model. We used \texttt{Keras} as our primary framework to create the transpiled neural network. So far our focus has been in developing the right architecture that can support a wide range of existing frameworks and in prototyping different custom features like \texttt{save()}, \texttt{explain()}, \texttt{nas()} and so forth.

\begin{table}[ht]
\centering
\begin{tabular}{ |p{3cm}||p{3cm}|p{3cm}|p{3cm}|p{3cm}|  }
%\begin{tabular}{ |m||m|m|m|m|  }
 \hline
 \multicolumn{4}{|c|}{Framework progress} \\
 \hline
 ML Square& Supported module&Algorithm&Status\\
 \hline
 \multirow{14}{*}{\texttt{.save()}}   & \multirow{12}{*}{\texttt{sklearn}}
 &Linear Regression&   \checkmark\\
 \cline{3-4}
 & &Ridge Regression&   \checkmark\\
 \cline{3-4}
 & &Lasso&   \checkmark\\
 \cline{3-4}
 & &ElasticNet&   \checkmark\\
 \cline{3-4}
 & &Support Vector Machines&   \checkmark\\
 \cline{3-4}
 & &Decision Tree Classifier&   \checkmark\\
 \cline{3-4}
 & &Linear Discriminant Analysis&   \small{\faFlask}\\
 \cline{3-4}
 & &Decision Tree Regressor&   \small{\faFlask}\\
 \cline{3-4}
 & &Multi-task Lasso&   \small{\faSpinner}\\
 \cline{3-4}
 & &Multitask ElasticNet&   \small{\faSpinner}\\
 \cline{3-4}
 & &Passive Aggressive Classifier&   \small{\faSpinner}\\
 \cline{3-4}
 & &Robust Regression&   \small{\faSpinner}\\
 \cline{2-4}
 & \texttt{IRT} & & \small{\faFlask}\\
 \cline{2-4}
 & \texttt{DeepCTR} & & \small{\faFlask}\\
 \cline{2-4}
 & \texttt{XgBoost} & & \small{\faSpinner}\\
 \cline{2-4}
 & \texttt{Surprise} & & \small{\faSpinner}\\
\hline
\multirow{3}{*}{\texttt{.explain()}} & \texttt{DeRT} & & \small{\faFlask}\\
\cline{2-4}
& \texttt{Interpret}& & \small{\faSpinner}\\
\cline{2-4}
& \texttt{Shapley}& & \small{\faSpinner}\\
\hline
\multirow{3}{*}{\texttt{.nas()}} & \texttt{Tune} & & \checkmark\\
\cline{2-4}
& \texttt{AutoKeras} & & \small{\faSpinner}\\
\hline
\texttt{.ci()} &  & & \small{\faFlask}\\
\hline
\end{tabular}
\caption{Progress of \texttt{mlsquare} features}
\label{progress}
\end{table}

The following list provides a gist of our ongoing research:
\begin{enumerate}
\item Extending \texttt{save()} to \texttt{XgBoost}, \texttt{IRT}\cite{irt}, \texttt{Surprise}\cite{Surprise} and \texttt{DeepCTR}\cite{DeepCTR}: Supporting these widely used frameworks will help us make deep learning techniques more accessible to a broader audience.
\item Adding support for Deep Random Trails(\texttt{DeRT}) in \texttt{explain()}: \texttt{DeRT} is an in-house explainability technique. We are currently working on on-boarding this tool to the \texttt{explain} method.
\item Adding support for \texttt{PyTorch}: As mentioned earlier, we currently use \texttt{Keras} to define the neural networks. We will soon be extending support to \texttt{PyTorch} as well.
\item Extending neural architecture search capabilities with \texttt{AutoKeras}\cite{AutoKeras}.
\item Bringing uncertainty quantification as a first class method on a model object.
\end{enumerate}

\subsection*{Contributing}
\texttt{mlsquare} is an open-source framework built by committed and enthusiastic volunteers. We strongly believe in the power of community-driven solutions. If you find \texttt{mlsquare}'s mission exciting, you can contribute to the framework in the following ways -

\begin{enumerate}
    \item Refer to this \href{https://github.com/mlsquare/mlsquare/blob/master/docs/developer.rst}{link} to add new algorithms to \texttt{dope} that you might find be useful to other developers.
    \item Please reach out to us at \texttt{info@mlsquare.org}, if you are interested in  any of the ongoing research or yet-to-begin projects, marked as [\small{\faFlask}] and [\small{\faSpinner}], respectively in Table \ref{progress}.
    \item This \href{https://docs.google.com/spreadsheets/d/1E5jcq2w42gN8bMIaeaRJpAdhgSVN-2XDJ_YTHe4qfwY/edit?usp=sharing}{spreadsheet} mentioned in \textit{Section} \ref{results section} is an internal tool used by \texttt{mlsquare} to keep a track of the performance of every algorithm available on the framework. We would like to add more datasets and create an openly accessible API to execute training and inference sessions for anyone interested in contributing. If you know of an interesting dataset or is interested in extending this tool, please do let us know by dropping an email at \texttt{info@mlsquare.org}.
    \item We are open to contributions at all levels(from documentation to architectural design inputs). To suggest changes to the framework, you can raise an issue on our GitHub \href{https://github.com/mlsquare/mlsquare}{repository}.
\end{enumerate}
\section{Conclusion}
In this paper, the need for democratizing AI, and several of its facets are introduced.  An extensible \texttt{Python} framework is proposed towards that end. Details of different components of the framework and their responsibilities are discussed. The framework currently provides support for porting a subset of \texttt{sklearn} models to their approximately faithful Deep Neural Network counterparts, represented in \texttt{ONNX} format. An important take-away is that, Deep Learning architectures can be constrained to reflect well-understood modeling paradigms. It debunks a common misconception that Deep Learning requires big data. In order to democratize the very creation of this framework, and bring transparency into the process, we have created a leader board to access the quality metrics of the semantic maps. Several enhancements and extensions are in pipeline. A process is to put in place to allow community to contribute.

\bibliographystyle{unsrt} 
\bibliography{references}  %%% Remove comment to use the external .bib file (using bibtex).
%%% and comment out the ``thebibliography'' section.
%\printbibliography

%%% Comment out this section when you \bibliography{references} is enabled.

\end{document}